\documentclass{article}
\usepackage{frExamplee}
\usepackage{graphicx}
\usepackage{apalike}
\usepackage{setspace}
\usepackage{amsmath,bm} 

\usepackage{wrapfig}

\usepackage{siunitx}

\usepackage{caption}
\usepackage{subcaption}

\usepackage{changes}

\usepackage{url}

\title{MixNet: Structured Deep Neural Motion Prediction for Autonomous Racing}

\author{
Phillip Karle$^{1}$\thanks{corresponding author}, Ferenc Török$^{1}$, Maximilian Geisslinger$^{1}$, Markus Lienkamp\\
Institute of Automotive Technology (FTM)\\
Technical University of Munich\\
Garching, Germany
}

\begin{document}

\maketitle

\footnotetext[1]{phillip.karle@tum.de, ferenc.toeroek@tum.de, maximilian.geisslinger@tum.de}


\begin{abstract}
Reliably predicting the motion of contestant vehicles surrounding an autonomous racecar is crucial for effective and performant planning. Although highly expressive, deep neural networks are black-box models, making their usage challenging in safety-critical applications, such as autonomous driving. In this paper, we introduce a structured way of forecasting the movement of opposing racecars with deep neural networks. The resulting set of possible output trajectories is constrained. Hence quality guarantees about the prediction can be given. We report the performance of the model by evaluating it together with an LSTM-based encoder-decoder architecture on data acquired from high-fidelity Hardware-in-the-Loop simulations. The proposed approach outperforms the baseline regarding the prediction accuracy but still fulfills the quality guarantees. Thus, a robust real-world application of the model is proven. The presented model was deployed on the racecar of the Technical University of Munich for the Indy Autonomous Challenge 2021. The code used in this research is available as open-source software at \url{www.github.com/TUMFTM/MixNet}.
\end{abstract}


\section{Introduction} \label{section:introduction}

Developing reliable autonomous vehicles has various purposes, including safer, more relaxing, and more efficient traveling. To fuel innovation in the field, autonomous vehicle competitions such as the DARPA Grand Challenge~\cite{darpa}, Formula Student\footnote[2]{https://www.imeche.org/events/formula-student/team-information/fs-ai} and Roborace\footnote[3]{https://roborace.com/} have taken place. There is, however, an aspect of autonomous racing that has not been covered before: full-scale multi-vehicle racing against other competitors. The Indy Autonomous Challenge (IAC) and its successor, the Autonomous Challenge at the CES 2022\footnote[4]{https://www.indyautonomouschallenge.com/} (AC@CES) were meant to take this enormous next step. In the competition, the teams were provided with the same hardware and developed their own autonomous software stacks. Wheel-to-wheel racing poses serious challenges, which are also present in everyday road traffic. To operate safely and efficiently in a dynamic environment, the vehicle has to predict the future motion of other dynamic objects. Usually, the methods tackling this problem take the motion histories of the objects together with some environmental information and make predictions conditioned on these. Several solutions use the current estimated state of the vehicle and extrapolate it using a physics-based model \cite{ammoun2009real,xie2017vehicle,deo2018would,schubert2008comparison}. Although these methods are appealing due to their transparency and small computational demands, the predictions become less reliable in the long term. Other approaches, which we will cover in detail, use machine learning techniques to match the motions of the objects to motion or behavior patterns and forecast accordingly. These methods often provide long-term predictions up to \SI{8}{\second} and are more expressive. On the other hand, the approximators used are often black-box models and hence performance or quality guarantees are hard to give.

\begin{figure}[b!]
    \centering
    \includegraphics[width=1.0\textwidth]{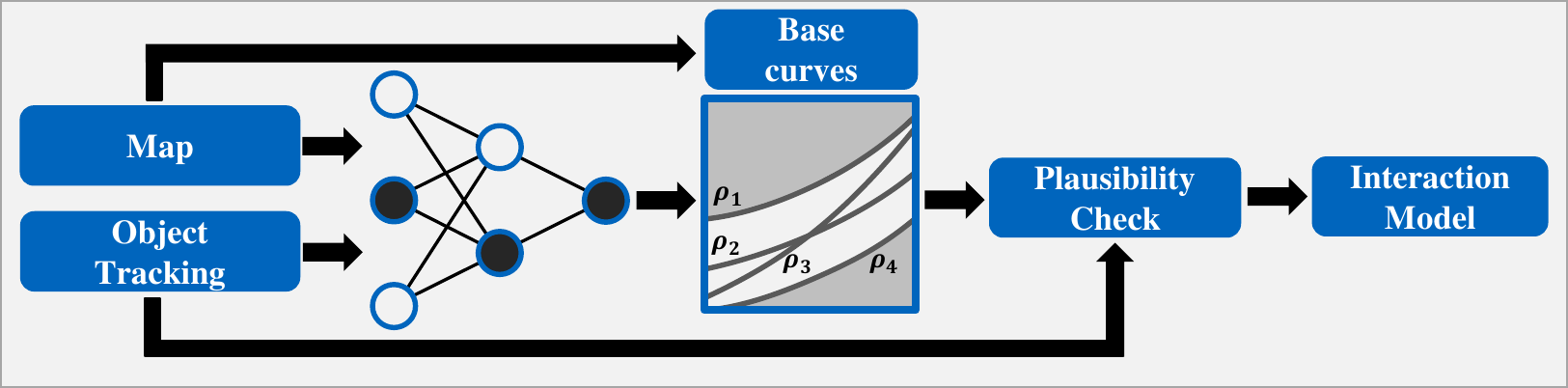}
    \caption{Overview of the MixNet prediction module. It combines a comprehensive, learned scenario understanding by means of an RNN-encoder and semantic knowledge to constrain the output by base curves extracted from the track map.}
    \label{fig:moduleoverview}
\end{figure}

\begin{figure}
	\centering
	\includegraphics[clip, trim=0.6cm 0.4cm 0.5cm 0.0cm 1.0cm,width=1.0\textwidth]{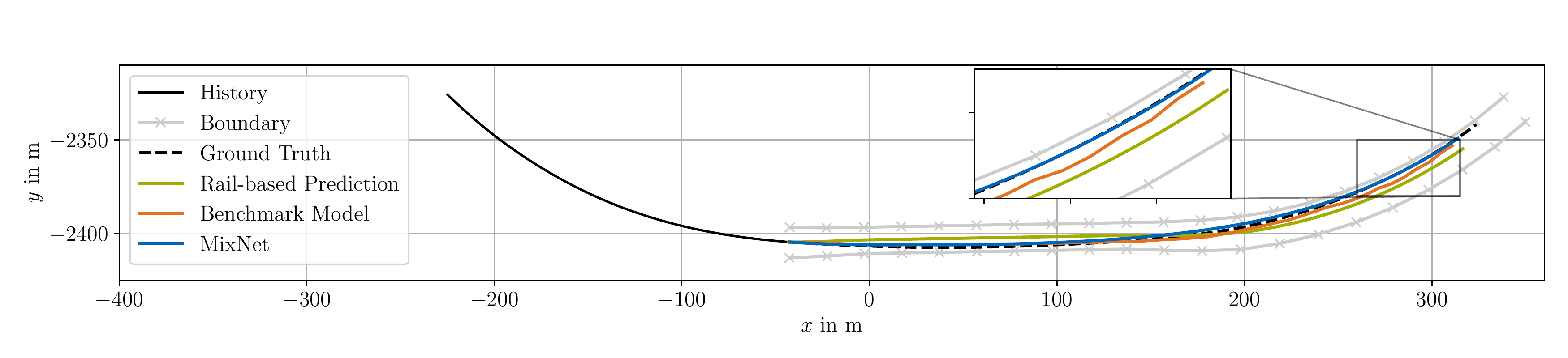}
	\caption{Exemplary data sample. The inputs to the MixNet, which are object history and sampled track boundaries, and the different prediction modes of a rail-based approach, the benchmark model and the MixNet are shown. MixNet combines the best of two worlds: the superposition of continuous base curves and the comprehensive, learned scenario understanding.}
	\label{fig:netinput}
\end{figure}

The prediction algorithm developed for the race has to fulfill some prerequisites: 1) it should output a single trajectory for each object; 2) the prediction horizon should be \SI{5}{\second} long; 3) the available time window for calculation is \SI{20}{\milli\second} with a single CPU core of computing capacity available; 4) it should be robust against noise in the inputs; 5) The output should always be smooth and should lie inside the racetrack. The available input information consists of the motion histories of the objects and the map of the racetrack. Our approach to fulfill these requirements is shown in Figure \ref{fig:moduleoverview}. We propose a structured deep neural motion forecasting model. The approach, although deep learning-based, creates trajectory prediction in a structured way. The object's past movement is encoded by Recurrent Neural Networks (RNNs) and transferred into a latent space. In contrast to common approaches, which determine the future trajectory in the decoder step entirely by deep neural network, MixNet calculates weighting parameters for superposition of dynamically feasible base curves derived from the track. By this, the MixNet is capable of providing strong quality guarantees for its predictions, which enables real-world application in a broad range of scenarios. An illustrative example is given in Figure \ref{fig:netinput}. MixNet outputs a prediction based on the encoded scenario understanding with high accuracy together with a smooth trajectory shape. In comparison, the benchmark model, which has the same encoder architecture, i.e., incorporates the same scenario understanding, but determines the prediction by means of a LSTM-based decoder cannot guarantee a stable output, so especially on a long prediction horizon the predicted trajectory gets noisy. Besides that, our approach comprises the feature of fusing external velocity information and overriding the calculated velocity profile. By this, the model can be further constrained, which are essential feature for a safe real-world application and useful to apply the model to a new Operational Design Domain (ODD). Besides that, an implemented fuzzy logic that models overtaking maneuvers improves the interaction-awareness to resolve non-feasible colliding trajectory predictions.

The structure of the paper is as follows. In Section \ref{section:related_work} we summarize the main research directions in the field of motion prediction for autonomous driving and discuss the applicability for our purpose regarding the aforementioned prerequisites. In Section \ref{section:method} we introduce our approach of the MixNet and additional features that cover the velocity fusion, safety checks and interaction-awareness. Besides that, we present our recorded dataset and the training procedure. Then in Section \ref{section:experiments} we evaluate its performance compared to the baseline model and analyze its robustness against noisy input and the superpositioning weights. In Section \ref{section:future_work} we discuss future research directions based on the shown results. Finally, in Section \ref{section:conclusion}, we conclude our work and outline the scope of the paper.


\section{Related Work} \label{section:related_work}
To give an overview of the state of the art in the field of motion prediction, we structure this section as proposed in \cite{karle2021review}. Accordingly, the methods are categorized based on the related investigation methodology, how to describe the object's motion, into physics-based, pattern-based, and planning-based approaches. After introducing the methods for motion prediction in general, we discuss the applicability for autonomous racing in the last subsection. For further details about the overall state of the art of software for autonomous vehicle racing the reader is referred to \cite{betz2022review}.

\subsection{Physics-Based Approaches} \label{subsec:physics_based_approaches}

In non-interactive scenarios with independent vehicle behavior and for short-term prediction up to \SI{2}{\second}, the application of kinematics- or dynamics-based vehicle models is a suitable choice \cite{thrun2006probabilistic}. Most commonly, deterministic or probabilistic kinematic models are propagated forward using a constant input assumption. One can choose various longitudinal and turning-related signals to be the input. Hence models with constant velocity, acceleration, turning radius and steering angle or the combination of these can be obtained. Schubert et al. \cite{schubert2008comparison} provide a review of these models and conclude that Constant Turning Radius and Acceleration (CTRA) models provide the best compromise between prediction accuracy and computational demands. To incorporate uncertainty information, Bayesian Filters, especially Kálmán Filters \cite{kalman1960new} are commonly used. If the linearity assumption of the Kálmán Filter does not hold, the Extended (EKF) or Unscented (UKF) versions are applied. By means of an Interaction Multiple Model (IMM) various kinematic models can be combined based on heuristics to improve the expressiveness in heterogeneous scenarios \cite{barrios2011improving}.\\
Another physics-based approach are reachable set predictions \cite{Koschi2017}, which utilize the set of physically possible behaviors. Thus all possible trajectories within the dynamic limits are determined, which are covered by a convex hull. Considering only the trajectories allowed by the traffic rules can limit the solution space accordingly and allows the use in a real vehicle for online verification \cite{Pek2020}. For the use in autonomous racing, these types of predictions are too conservative for the long-term prediction due to the wide range of driving dynamics of the racing vehicle and the lack of explicit rules as in road traffic. The application of the online verification concept in a supervisor module with the use case of autonomous racing is shown in \cite{stahl2020online}.\\
In general, physics-based methods are computationally cheap and their operation is transparent and well studied, which makes them appealing in safety-critical application domains such as autonomous driving. The main limitation arises from the simplified input assumption. This results in fairly accurate predictions on the short horizon. However, the prediction gets quickly outdated in the long term (\SI{>2}{\second}) as the assumption of constant movement does not hold anymore. Hence, these methods are often combined with other approaches which tend to produce more reliable long-term predictions \cite{xie2017vehicle,deo2018would}.

\subsection{Pattern-Based Approaches} \label{subsec:Pattern_based_approaches}

Pattern-based approaches build on the idea of taking observations of an object, matching it to a pattern, and carrying out the prediction based on it. The pattern assigned to a vehicle can be handcrafted or learned. Furthermore, patterns can be learned in physical by clustering data points or in an abstract space such as the hidden representations of Encoder-decoder models. Most of these methods are data-based approaches, hence the need for sufficiently rich datasets is inherent.

When using handcrafted patterns, the motion of an object is assigned to one of the predefined maneuver classes. Then, the output prediction is constructed considering the assigned maneuver type and usually involves the usage of prototypes. The classification can be carried out based on heuristics \cite{houenou2013vehicle} or ML models, such as Support Vector Machines (SVMs) \cite{mandalia2005using}, Hidden Markov Models (HMMs) \cite{yuan2018lane,li2019generic,bicego2004similarity} or RNNs \cite{khosroshahi2016surround}. HMMs and RNNs are commonly used due to their inherent capability of interpreting the temporal evolution of a motion history. Instead of using handcrafted patterns, it is also possible to learn clusters from data. Afterwards, a single prototype trajectory \cite{morris2011trajectory} or a probabilistic representation \cite{kruger2019probabilistic} is obtained for each cluster. During inference time, the output is constructed by classifying the given scenario into a specific cluster and applying the respective representant.

Encoder-decoder neural network architectures have recently shown huge potential for motion prediction tasks by dominating the leaderboards of various prediction challenges on public real-world datasets \cite{chang2019argoverse,caesar2020nuscenes}. For the use case of motion prediction, the encoder creates a latent summary about the motion history of an object and the environment information, which serves as an abstract pattern of the scenario. Conditioned on this abstract representation, the decoder determines the future movement of the object. The encoder and the decoder models can both be based on Convolutional Neural Networks \cite{harmening2020deep,ridel2020scene}, RNNs \cite{sadeghian2018car,li2019conditional} or even Convolutional-RNNs \cite{xingjian2015convolutional,ridel2020scene}.

Attention mechanisms are also commonly used, to allow the network to focus on the relevant parts of the input \cite{sadeghian2018car,li2019conditional} or to take interactions between vehicles into account \cite{mercat2020multi}. Alternatively, Graph Neural Networks (GNNs) can be used to model interactions in a non-euclidean space. A learned graph representation of the scene is applied to model individual interaction between the single agents. The application on public real-world datasets shows a significant improvement in prediction performance \cite{salzmann2020trajectron,zhao2020gisnet}. Another advantage of graph-based prediction models is the more efficient representation in contrast to grid-based approaches, which leads to a reduced calculation time \cite{gao2020vector}. The bottleneck to provide a sufficiently rich dataset to train a performant algorithm is mitigated in terms of Encoder-decoder architectures due to the fact that no labeled data is needed to create a dataset. The reason for this is that the observation of an object's movement serves as ground truth for output of the prediction algorithm. Thus, the trajectories can simply be split into history and ground truth prediction parts and used in the self-supervised learning setup \cite{geisslinger2021watch}.

\subsection{Planning-Based Approaches} \label{subsec:Planning_based_approaches}

Planning-based approaches consider objects to be rational agents acting in an environment according to their hidden policies in order to reach their goals. The basis of planning-based approaches is the Markov Decision Process (MDP). The approaches usually differ from each other in approximating different parts of the MDP. One approach is to derive handcrafted cost functions to model the agent's behavior \cite{rudenko2018joint}. However, the creation of a comprehensive cost function and rule set requires dedicated knowledge and complex traffic scenarios are challenging to represent. Alternatively, the driving behavior can be learned from data, which refers to the field of learning from demonstration. On the one hand, it is possible to apply Inverse Reinforcement Learning, which aims to derive a cost function that fits the observed expert behavior \cite{sun2018probabilistic,deo2020trajectory,fernando2020deep}. On the other hand, in the case of Imitation Learning the policy is directly learned from the observed data, i.e., the observation is directly related to a specific behavior \cite{bansal2019chauffeur}. To enhance the robustness of the learned policy generative methods are used, one common approach is Generative Adversarial Imitational Learning \cite{kuefler2017imitating}.

Both pattern-based and planning-based approaches are usually capable of producing more reliable long-term predictions compared to physics-based methods because more comprehensive features can be inputted to the model to consider interaction between different road users and map constraints. Their shortcomings come from the fact that these methods are primarily data-based and rely on learned policies or patterns. Hence, their performance highly depends on the underlying dataset, which has to represent the ODD sufficiently. Besides the amount of data, the balance of scenario types influences the prediction performance in edge cases such as safety-critical situations. So, the under-representation of these scenarios directly impacts the applicability. The application in combination with safeguarding methods is also challenging as data-based methods lack of explainability. Hence their behavior can not be supervised properly.

\subsection{Applicability for Autonomous Racing}
In the following the state of the art is evaluated regarding the applicability for autonomous racing. In our case we focus on the Indy Autonomous Challenge (IAC), one of the most advanced autonomous racing series, which is the target for the presented approach. In the inaugural edition the IAC was held on the Indianapolis Motor Speedway (IMS) and the Las Vegas Motor Speedway (LVMS), which are both oval circuits with additional simulation races in advance. The circuits do not offer any lanes to define lane-keeping or lane-changing maneuvers. Also, overtaking can have many different forms because the lane layout does not constrain it as in normal road traffic and it is assumed to be non-cooperative. Besides these constraints resulting from the race format, it also has to be considered that no public race dataset is available and the execution time on fixed computational resources has to fulfill real-time requirements.
Due to the overall software concept \cite{tumiacpaper}, it is required to provide a prediction horizon of \SI{5}{\second}, which is another important constraint. In consideration of missing traffic lanes, the prediction length does not allow the usage of purely physics-based methods as constant movement assumptions do not hold and the set of dynamically reachable states gets too large. The factors of missing lane information and data discourage the application of classification or clustering-based approaches. The application of planning-based algorithms is questionable because these algorithms require even more comprehensive data to derive the expert behavior properly and, especially in terms of IL, the robustness towards outliers is not given.

The design factors that motivate our approach, the MixNet, is to combine data-based encoding with dynamically feasible superposition of base curves are the real-time capability and the need for robustness and performance guarantees. By means of the Encoder network we can extract patterns from observation. These patterns cover comprehensive object behavior, so complex scenarios can be modeled. The superposition of base curves in the decoder constrains the possible prediction to lie inside the race track and it guarantees robustness in case of outliers in the input. Thus, the applicability can be enhanced significantly. However, the superposition of base curves is still flexible enough to output a high prediction accuracy as the evaluation (section \ref{section:experiments}) shows. To model interactions and to output collision-free predictions a rule-based fuzzy logic is implemented which can be applied to the MixNet's output.


\section{Method} \label{section:method}

In this section, we introduce MixNet, our encoder-decoder neural network architecture for motion prediction. Besides the network architecture, we additionally describe how to incorporate external velocity information, trigger safety interventions and model interactions between objects. Finally, we describe the data mining and training procedure.

\subsection{Network Architecture}

\begin{figure}[b]
    \centering
    \includegraphics[width=\textwidth]{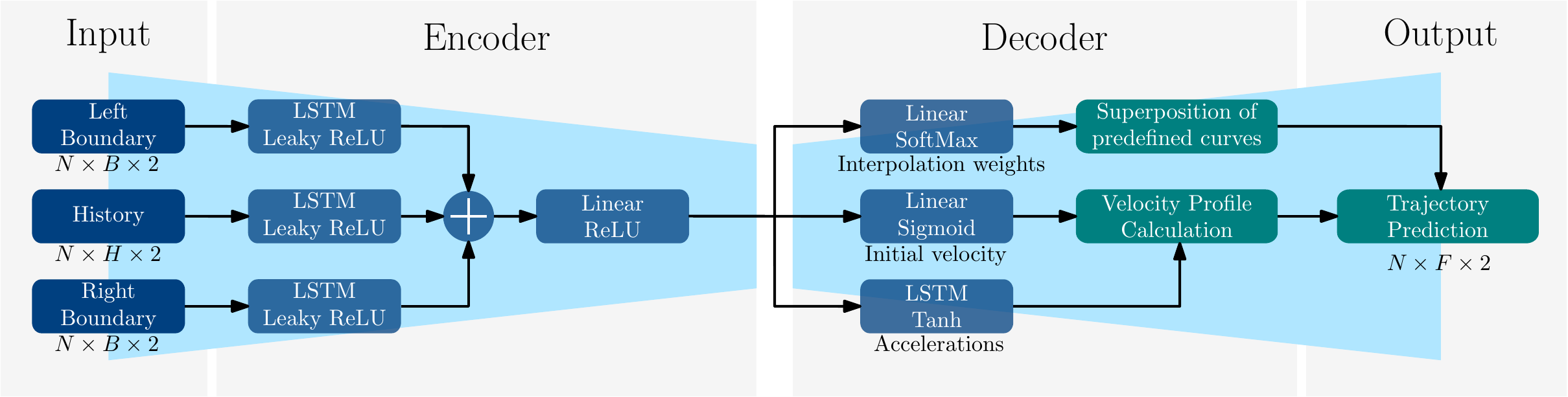}
	\caption{The architecture of MixNet. The prediction is composed of a path prediction by superposition of base curves and the prediction of an acceleration profile to apply a piece-wise linear velocity profile. Inputs to the network are $H = 30$ past 2D-positions of $N$ objects and the related left and right track boundaries in driving direction in vector representation. Output is the trajectory prediction of $N$ objects in 2D with a horizon of $F = 50$ steps.}
	\label{fig:MixNet_architecture}
\end{figure}

The proposed network architecture of the MixNet is shown in Figure \ref{fig:MixNet_architecture}. The LSTM layers in the encoder create a latent summary through encoding the object motion histories. The inputs to the network are the $H = 30$ historic 2D-positions up to \SI{3}{\second} in the past, sampled with $f=$ \SI{10}{\hertz}. Besides that, the relevant map information, which are the left and right track boundaries starting from the current object's position, equidistantly sampled in vector representation, is inputted to the network. Considering the expected racing speed and prediction horizon, we sample the boundaries up to a horizon of \SI{400}{\meter} with \SI{20}{\meter}. During inference $N$ objects are fed into the network batch wise. Figure \ref{fig:netinput} shows an exemplary input consisting of history and track boundaries. The hidden states of the encoding LSTMs are then concatenated and passed to a linear layer, which outputs the latent representation of the scenario. The decoder, the generative part of the model, creates a prediction conditioned on the latent summary. In contrast to other works, the future states are not the direct output of an LSTM network, but the forecast trajectory is generated systematically from known schemes. First of all, the trajectory is obtained by predicting a path and applying a velocity profile to it. Both of these components are generated in a constrained way. The path is created through superpositioning various predefined base curves according to weights predicted by the network. The velocity profile is piece-wise linear and is determined by predicting an initial velocity and five constant acceleration values: one for each second of the prediction horizon. The final output trajectory is obtained by resampling the path according to the velocity profile. In this way, the set of possible output trajectories is constrained by construction.

We use four base curves for superpositioning: the two track boundaries, a pre-computed minimal curvature raceline \cite{heilmeier2019minimum} and the centerline of the track. The curves are represented by discrete 2D-points of equal number. The points of the boundaries and the raceline are defined by their distances from the centerline points. Hence, the points of each curve at any index $i$ correspond to the same cross section on the track. Due to this fact, the superpositioned curve can be obtained as follows:
\begin{align}
\rho_{sup}(s) & = \sum_{c \in \mathcal{C}}\lambda_c \rho_c(s) \text{ with} \label{eq:1}\\
\lambda_c \in [0, 1] \; \; \forall c & \in \mathcal{C} \text{ and }
\sum_{c \in \mathcal{C}}\lambda_c = 1 \label{eq:2}
\end{align}
where $\rho_{sup}(s)$ is the superposed curve along the arc length $s$ of the track with base curves $\rho_c(s)$ and their corresponding weights $\lambda_c$ in the set of base curves $\mathcal{C}$. Due to the fact that all base curves are sampled along the same cross sections, $s$ is equal for all base curves. Equation \ref{eq:2} provides the constraints for the superpositioning weights, which result in curves that lie between the left and right boundaries. These constraints can easily be enforced by applying the \textit{softmax} activation to the output of the linear layer.

We emphasize that this method does not guarantee that a predicted trajectory starts exactly from the actual position of a vehicle, but a lateral offset can occur. This is due to the fact that the superposition weights by Equation \ref{eq:1} are constant along the path and the model is trained to output a prediction with the smallest overall error along the prediction horizon. Hence, a lateral offset at the beginning of the predicted path is possible. To remedy this issue, during the first second of the motion, we apply a plausibility check to the output of the network by means of a comparison between the current object position and the predicted path. Above a specified threshold, a correcting shift is triggered to ensure a consistent transition between the current position and the predicted path. Especially in transient scenarios, this characteristic is beneficial: If the current motion of the vehicle is strongly transversal, the network learns to predict a path that fits the later parts of the ground truth better by sacrificing accuracy at the beginning of the horizon. With the application of the correction shift, the consistency of the predicted motion is secured, so an additional quality guarantee can be given. During inference, if an adjustment of the the first part of the trajectory is triggered to connect the actual position to the prediction path, trajectories similar to lane-change maneuvers can be obtained.

\subsection{Velocity Information Fusion} \label{subsec:velocityinformationfusion}

A further advantage of the modular trajectory prediction is that it can also incorporate velocity information from another source to fuse it with the outputted path of the network.
The proposed implementation of the MixNet offers two possibilities for incorporating external velocity information. First, one can take the complete velocity profile from an external source and use it instead of the one predicted by MixNet. The other possibility is to predict the piece-wise constant accelerations with MixNet but use the external velocity information only to determine the current velocity as initialization.

A reliable source of such information is object tracking which contains the filtered states of the surrounding vehicles. While the current velocity can be extracted from the tracked object's state independently from the ODD a complete velocity profile underlies specific assumptions. For our use case of autonomous racing on a closed track, we propose to determine a complete velocity profile along the whole prediction horizon by forward propagating the tracked object's state by means of the underlying state-space model. However, this assumption is limited to scenarios with objects at race speed. Experimental evaluation shows that the combination of the tracked initial velocity and the piece-wise constant acceleration predicted by the MixNet provides the most accurate velocity profile predictions. Hence, similar to the path prediction, the combination of an external constraint by means of the initial velocity and the scenario-aware data-based acceleration prediction performs best.

\subsection{Safety Override Possibility}
Predicting accurately at the beginning of the horizon is a safety-relevant issue. On the other hand, large inaccuracies towards the end of the prediction horizon lead to inefficient planning and hence bad performance during the race. Due to the constant superposition weights, this trade-off especially occurs at the turn-in point and during overtaking maneuvers. To solve this conflict, our prediction method recognizes and overrides dynamically infeasible predictions, that is, predictions which have a large lateral offset at the beginning of the horizon. Accordingly, our goal is to probabilistically identify the cases where this happens. Our measure of probability for having generated an initially highly inaccurate forecast is based on the raw path prediction output of the MixNet and the actual state of the vehicle. As stated before, without adjustment, the predicted path does not necessarily start at the actual position of the object. Thus, if the predicted path lies too far from the vehicle considering its actual position and orientation, the prediction is identified as invalid. In this case, we override it with a prediction approach, which we call \textit{rail-based} prediction. This approach is derived from the tracked object state and offers a high robustness and guarantees kinematic consistency with the current object's state, which is of high importance as fallback option. The rail-based prediction is composed out of a separate path and velocity profile. The path is sampled starting from the current object position in parallel to the track boundaries. The velocity profile is determined by forward propagation of the state-space model as described in \ref{subsec:velocityinformationfusion}. An exemplary prediction by means of the rail-based approach is shown in Figure \ref{fig:netinput}. It can be seen that the rail-based prediction is consistent and accurate at the beginning of the prediction horizon, but has a high lateral error on the long-term horizon.

\subsection{Interaction Modeling}
As it can be seen in Figure \ref{fig:MixNet_architecture} no information about surrounding race cars is fed into the neural network. Hence, there is no explicit interaction-awareness given by the model. Although this assumption holds for many race scenarios with cars following their raceline, interactive scenarios such as overtaking or blocking, which are highly interactive, require additional modeling. In the literature, these interactions are modeled using game theory \cite{smirnov2021} or learned by a prediction network \cite{Deo2018}. These approaches have the disadvantage that they either require a lot of computing time due to iterations or require a vast amount of relevant data with interactions. 

Therefore, we propose a two-step approach for trajectory prediction on the racetrack. This first predicts each vehicle by itself and then adjusts the predicted trajectories based on interactions in a second step. In doing so, we take advantage of the set of rules in motorsports. Similar to other racing series such as Formula 1, for safety reasons the movement options of an overtaken vehicle are restricted by the rule set. In the Indy Autonomous Challenge, the vehicle in front is even forced to "hold its raceline" \cite{iacrules2019}. In other words, it may not block the overtaking vehicle or initiate other unpredictable maneuvers.

Our approach is to predict each vehicle individually, including the ego-vehicle, as a first step. Subsequently, all existing predicted trajectories of the different vehicles are examined for collisions. If no collision is detected, we assume that the influence of the interaction is minor and no adjustment of the trajectories is necessary. However, if at least one collision is detected, we do adapt the trajectories to account for interactions. For this, we examine the race order for each collision and adjust the predicted trajectory of the rear vehicle, since the front vehicle is not allowed to adjust its behavior according to the rules.

To adjust the predicted trajectory, a high-level decision is first made: the faster rear vehicle will either overtake on the left, overtake on the right or not overtake at all and stay behind the front vehicle. This high-level decision is made by a fuzzy logic that takes into account the absolute and relative positions as well as the velocities of the participants. According to the decision, the predicted trajectory is adjusted laterally (in case of an overtake) or longitudinally (in case of no overtake). The adjusted trajectories do not necessarily have to be collision-free, since new collisions with other predicted trajectories may occur. Therefore, the procedure can be repeated as often as necessary until all predictions are resolved collision-free or a termination criterion occurs. In our application, it has proven to be useful in a second iteration to ensure only that the ego vehicle is collision-free with vehicles in the rear. Otherwise it could happen that the trajectory planning tries to avoid this collision and even makes room for an overtaking vehicle, which would be contrary to winning the race.

\subsection{Dataset} \label{subsec:dataset}
An essential part of successful Machine Learning applications is a rich dataset, which covers a diverse set of scenarios expected during inference. However, since this race is the first of its kind, building a dataset from previous races is not possible. Also, there is no public racing dataset available. In this respect, one of the key challenges to be solved is building a dataset for training our neural network approach.

Real-life data was not available until the last weeks before the race itself since real-world tests on the IMS track only took place then. Hence, training data had to be acquired through simulating our software pipeline \cite{tumiacpaper} against itself on a high fidelity Hardware-in-the-Loop (HIL) simulator. The HIL-simulator is able to simulate the full autonomous software stack of up to ten agents in real-time with the same interfaces and callback functions as on the real vehicle. The only constraint is a simplified perception input to reduce the computation load, i.e., the perception pipelines are bypassed with a synthetic object list generator that is input to the tracking module. However, this generator comprises various features to imitate real perception behavior such as limited sensor range and field-of-view, addition of Gaussian and normal noise, variation of measured objects states and the simulation of false positives and false negatives. Multi-vehicle races on the HIL-simulator and data from the official simulation race of the IAC, both with highly interactive and complex scenarios, are the basis of our dataset. The recorded logs contain the output of the tracking module, which are used to recreate the trajectories of each vehicle during a race. Using the tracking module output to build our dataset has the advantage that the training input distribution of tracked objects and tracking quality will be as close as possible to the expected input distribution during the race. Besides that, the synthetic object list generator is used to vary the perception quality with the aforementioned features to augment the dataset. We added Gaussian noise during the data generation process with mean $\mu = 0$ and different variance in longitudinal and lateral directions based on the evaluated perception and tracking performance.

From these recovered trajectories and the map of the racetrack, it is possible to generate a dataset for training the model. As it was stated before, the input of our model consists of the $(x, y)$ positions of the history trajectory and the track boundaries around and ahead of the vehicle. Before inputting these points to the model, they are transformed into a local coordinate system, which has its origin at the left boundary next to the current object position and is oriented with its $x$-axis tangential to the left boundary. With the process above we managed to recreate 1,569 trajectories from 226 races with in average 3.4 vehicles per race. From these trajectories, we created 358,025 input-output data pairs.

\subsection{Training}

The loss function we defined for training has two terms, one for costing the fit of the interpolated curve, which relates to the error in lateral direction, and one for the velocity profile, which reflects the accuracy in the longitudinal direction. The loss $L_{\mathrm{path}}$ related to the lateral deviation of the interpolated curve $\bm{\hat{x}}$ from the ground truth $\bm{x}$ is costed with a Weighted Mean Square Error (WMSE) at the beginning of the curve as follows:

\begin{equation}
 L_{\mathrm{path}} = \frac{1}{F} \left( L_{\mathrm{wmse}} + L_{\mathrm{mse}} \right)
\end{equation}
\begin{equation}
L_{\mathrm{wmse}} = \sum_{i=1}^{k}(\bm{\hat{x_i}} - \bm{x_i})^2 \left(1+w \left( 1 -\frac{i-1}{k} \right) \right)
\end{equation}
\begin{equation}
L_{\mathrm{mse}} = \sum_{i=k+1}^{F}(\bm{\hat{x_i}} - \bm{x_i})^2
\end{equation}
\begin{equation}
\text{ with } w = 0.5, k = 10, F = 50
\end{equation}

The WMSE decreases linearly from the first prediction step with an additional weight $w$ along the weighting horizon $k$. The remaining prediction horizon is weighted with $1.0$, which corresponds to the conventional Mean Square Error (MSE). The weighting turned out to be beneficial to limit the lateral offset at the beginning of the prediction, even if an additional correction shift is applied at inference.
The loss of the velocity profile is also the MSE coming from comparing it to that of the ground truth velocity profile. The overall loss $L$ is then obtained from the path loss $L_{\mathrm{path}}$ and the velocity loss $L_{\mathrm{vel}}$ as follows:

\begin{equation}
    L = L_{\mathrm{path}} + \Delta t^2 \cdot L_{\mathrm{vel}}
\end{equation}

Where $\Delta t = 1 / f = \SI{0.1}{s}$ is the timestep size of the prediction in seconds. The multiplication of the velocity error with $\Delta t ^2$ comes from the following intuition: The velocity error $e_v$ results through integration in a $\Delta t \cdot e_v$ displacement error and a $\Delta t^2 \cdot e_v^2$ squared displacement error, if the MSE is applied. By considering the relation $e_v^2 \sim L_{\mathrm{vel}}$, it follows that multiplying the velocity loss term with $\Delta t^2$ is necessary to be able to add two loss terms of an identical unit, which is m\textsuperscript{2}.

The hyperparameters are optimized by Bayesian optimization \cite{nogueira2014bayes}. To train the network we use a learning rate of $5 \cdot 10^{-5}$ with a rate decay of $0.997$ per epoch. The $L_2$ weight regularization has a strength of $10^{-7}$ and we use a batch size of $128$. We train the model for $50$ epochs and take the model with the best validation for the evaluation presented in Section \ref{section:experiments}. The final net and training parameters are published in the open source code. 

In conclusion, the proposed method offers guarantees that the predicted trajectories are always smooth and lie inside the racetrack by means of the structured composition of the prediction. Besides that it is possible to fuse velocity information from other sources such as the state estimation to enhance the kinematic consistency of the predicted trajectory. Finally, it is possible to probabilistically detect predictions that are highly incorrect at the beginning of the horizon. In these cases, it is possible to override the predictions. For autonomous racing on a closed track, we propose to use a rail-based prediction, which outputs a constant velocity trajectory in parallel to the track boundaries starting from the current object's position.

\section{Experiments} \label{section:experiments}

In this section, we describe the test procedure, which comprises details about the test dataset and present a comprehensive evaluation of MixNet's prediction performance. The conducted experiments reveal the overall prediction performance and analyzes the model's robustness. Besides that, we investigate the composition of the base curves.

\subsection{Test procedure}

The set of predictions that are reachable through MixNet is constrained. This is exactly what allows for the quality guarantees mentioned before. To demonstrate that, despite this boundedness, MixNet is still flexible enough to output an accurate prediction, we compare its performance to a purely LSTM-based encoder-decoder architecture, which we define as benchmark model. The benchmark model's encoder architecture is identical to MixNet's. The difference lies in the decoder architecture, which is in case of the benchmark model also constructed with LSTM-layers. Thus, the model directly iteratively outputs a 2D trajectory prediction with a shape of $F\,\mathrm{x}\,2$ for $N$ objects. MixNet and the benchmark model have $198,214$ and $193,797$ trainable parameters respectively. We train both models on the same dataset which is described in subsection \ref{subsec:dataset}. For MixNet, we incorporate initial velocity information from object tracking into the velocity profile, but use the piece-wise constant acceleration outputted by the MixNet.

For reproducibility, we have recorded $10$ scenarios on our HIL simulator. These include interactive race scenarios with different numbers of vehicles with various speed limits. The recordings can be replayed identically to the pipelines using the two prediction models. For the recording of the interactive scenarios, we used MixNet as the prediction model to run the full software stack. We would like to emphasize that this does not induce any bias in the evaluation process as the respective planning behavior of each objects differs from the MixNet model behavior. We report the performances of the models by analyzing the absolute error distributions in the lateral and longitudinal directions with respect to the horizon length using the Mean Average Error (MAE), which is defined as follows:
\begin{equation}
\mathrm{MAE} = \frac{1}{F}\sum_{i=1}^{F}|\bm{\hat{x_i}} - \bm{x_i}|_2
\end{equation}

\subsection{Overall Prediction Performance}
\begin{figure}[b]
    \centering
	\includegraphics[width=0.49\textwidth]{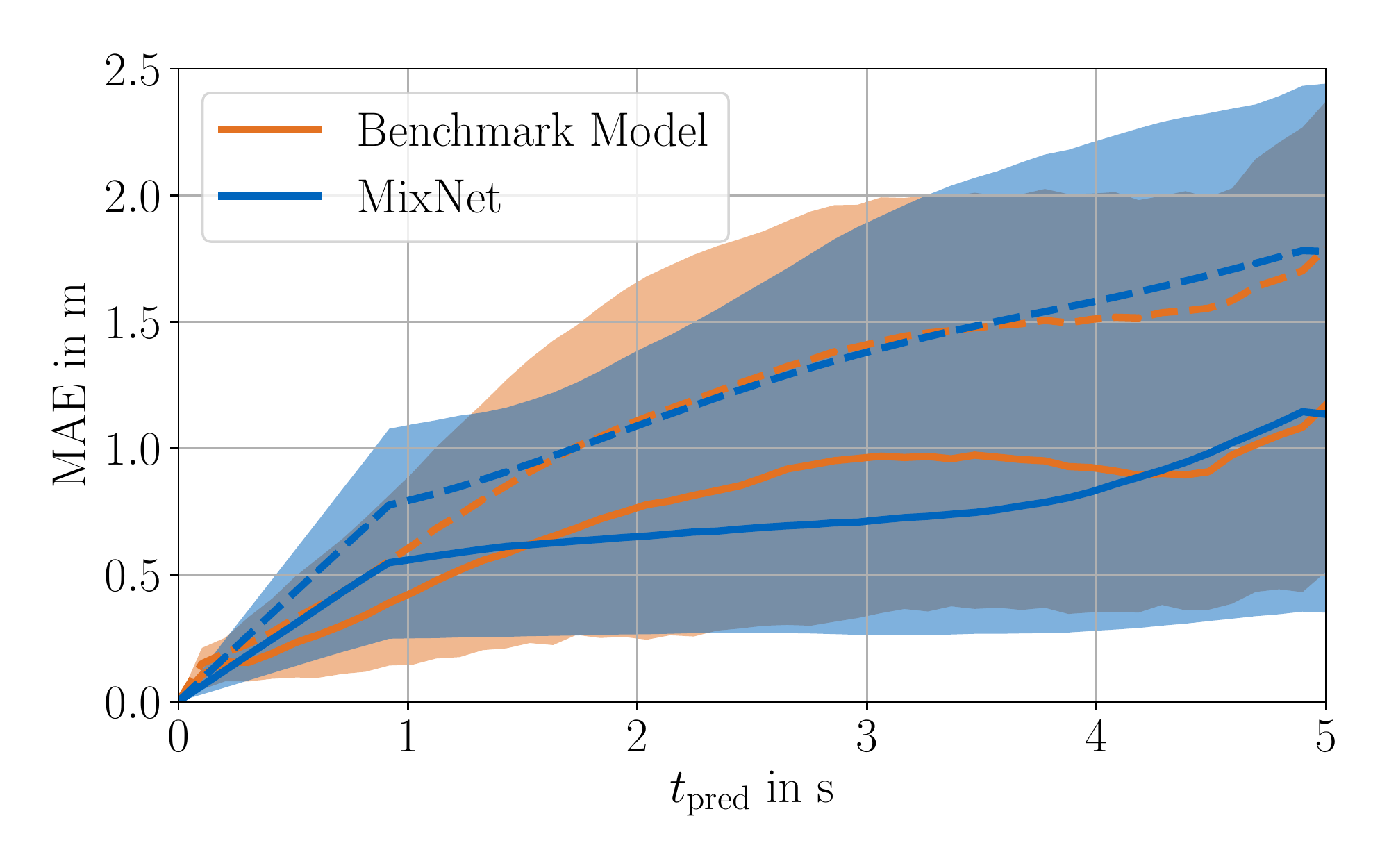}
	\includegraphics[width=0.49\textwidth]{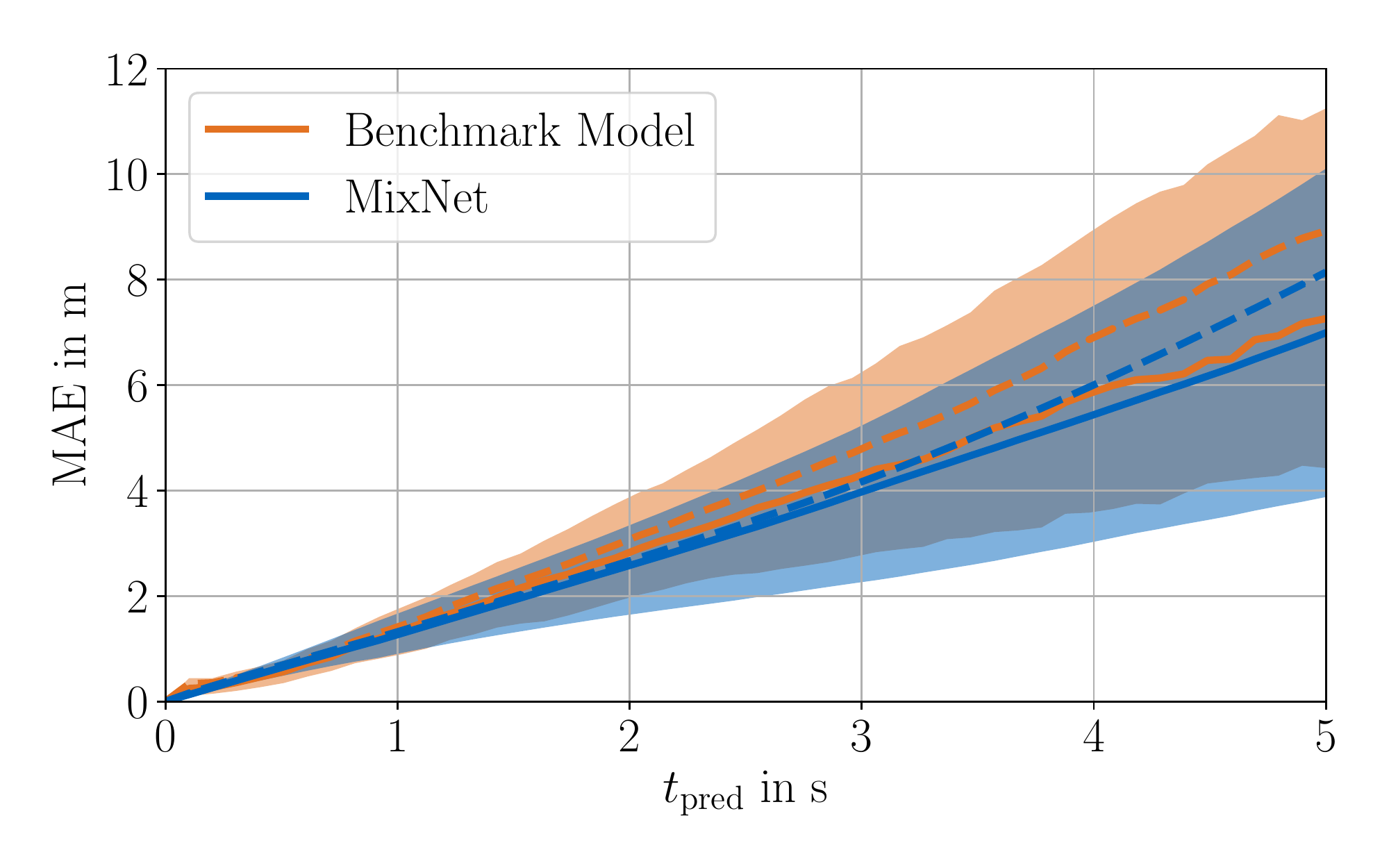}
	\caption{Lateral (left) and longitudinal (right) error distribution on the prediction horizon. The solid and dashed lines denote the median and mean errors respectively. The colored areas illustrate the range between the Q1 and Q3 quartiles.}
	\label{fig:errors_on_the_horizon}
\end{figure}

The overall MAE on the recorded interactive scenarios of MixNet and the benchmark model are \SI{4.91}{\meter} and \SI{5.36}{\meter} respectively. The reason why MixNet, although being constrained, outperforms the benchmark model becomes apparent when we look at the lateral and longitudinal error distributions on the prediction horizon illustrated in Figure \ref{fig:errors_on_the_horizon}. As it can be seen, the magnitude of errors in the lateral direction is very similar in the two cases. This result justifies the hypothesis that obtaining the prediction path by superpositioning our chosen base curves covers the set of possible trajectories properly. The superiority of MixNet in the overall error comes from the fact that it produces smaller errors in the longitudinal direction. Thus, it can be concluded that the combination of the initial velocity from the tracking module and the assumption of piece-wise constant acceleration models the longitudinal movement of the objects more accurate than the iteratively determined output of the LSTM-decoder.

\begin{figure}[h!]
     \begin{subfigure}[b]{0.48\textwidth}
         \centering
    	\includegraphics[width=0.45\textwidth]{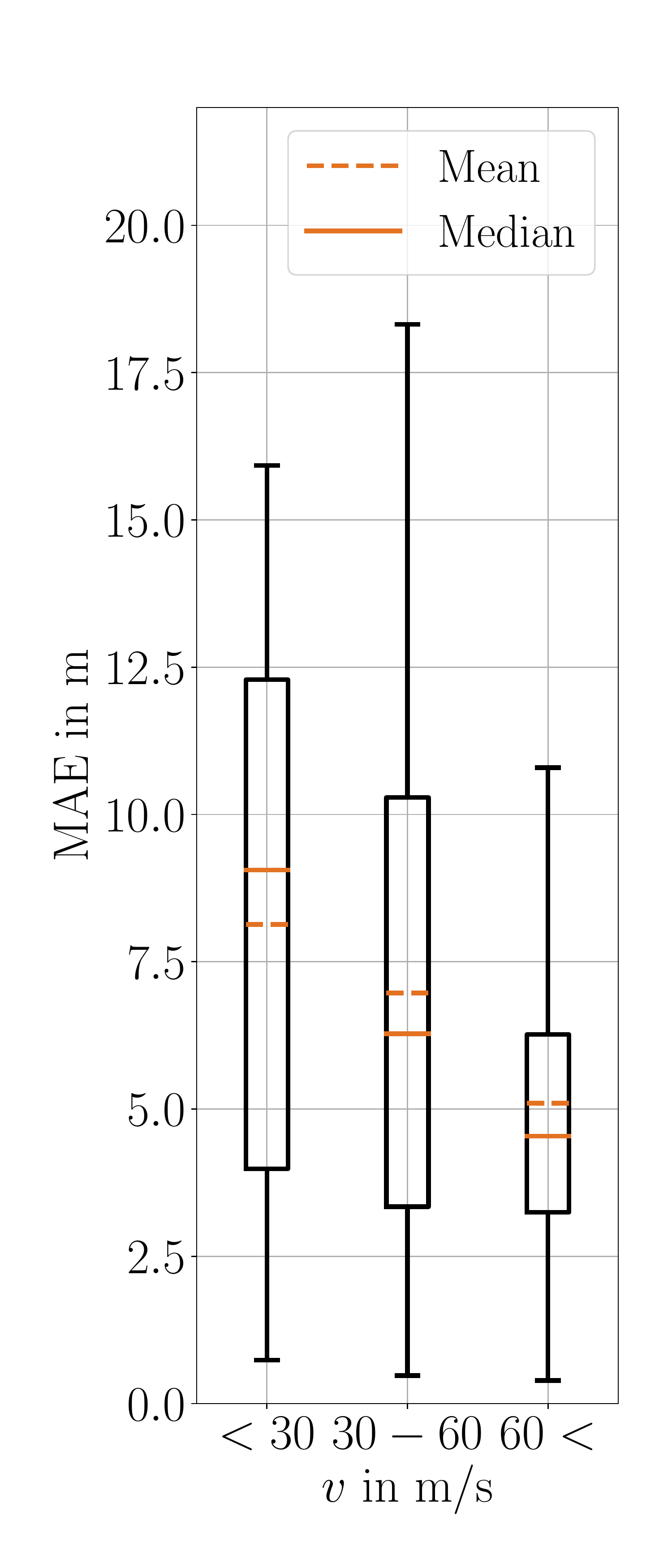}
    	\includegraphics[width=0.45\textwidth]{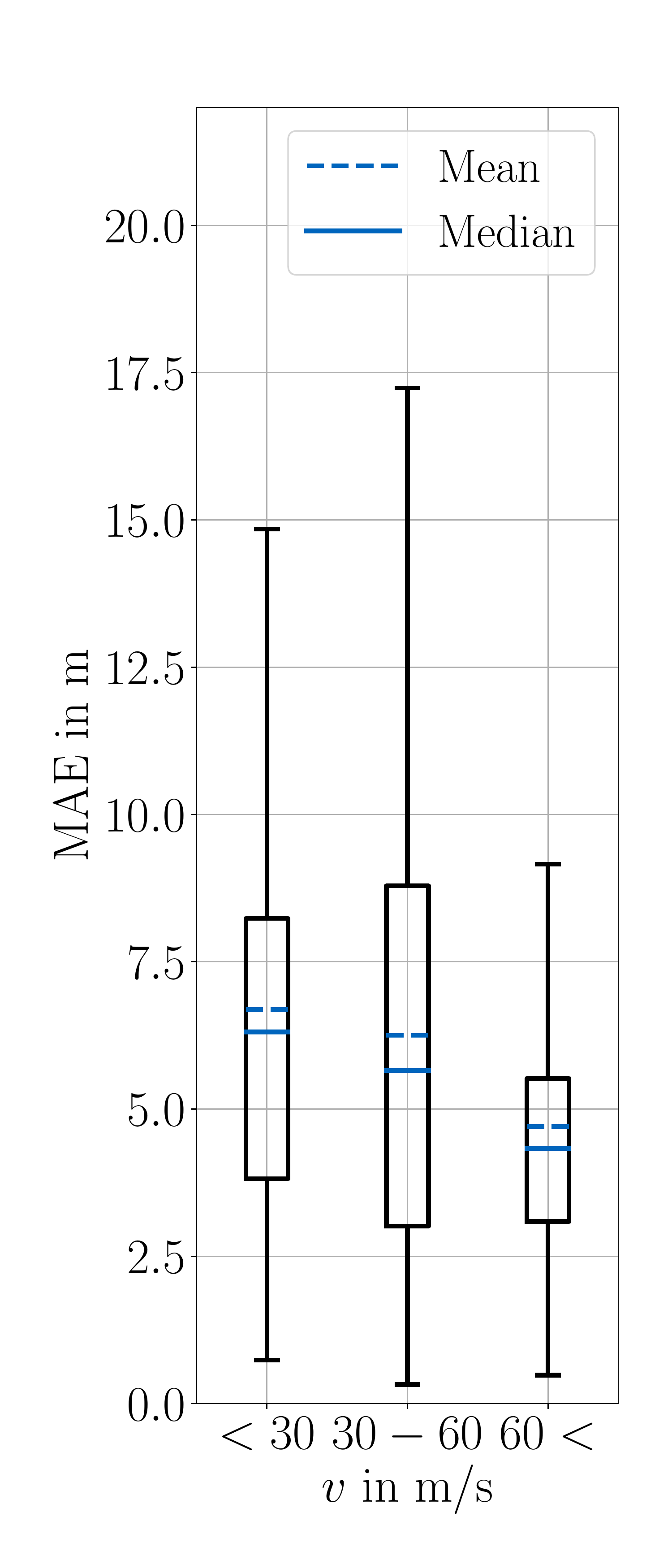}
        \caption{Object velocity.}
        \label{fig:errors_vs_velocity}
     \end{subfigure}
     \begin{subfigure}[b]{0.48\textwidth}
        \centering
    	\includegraphics[width=0.45\textwidth]{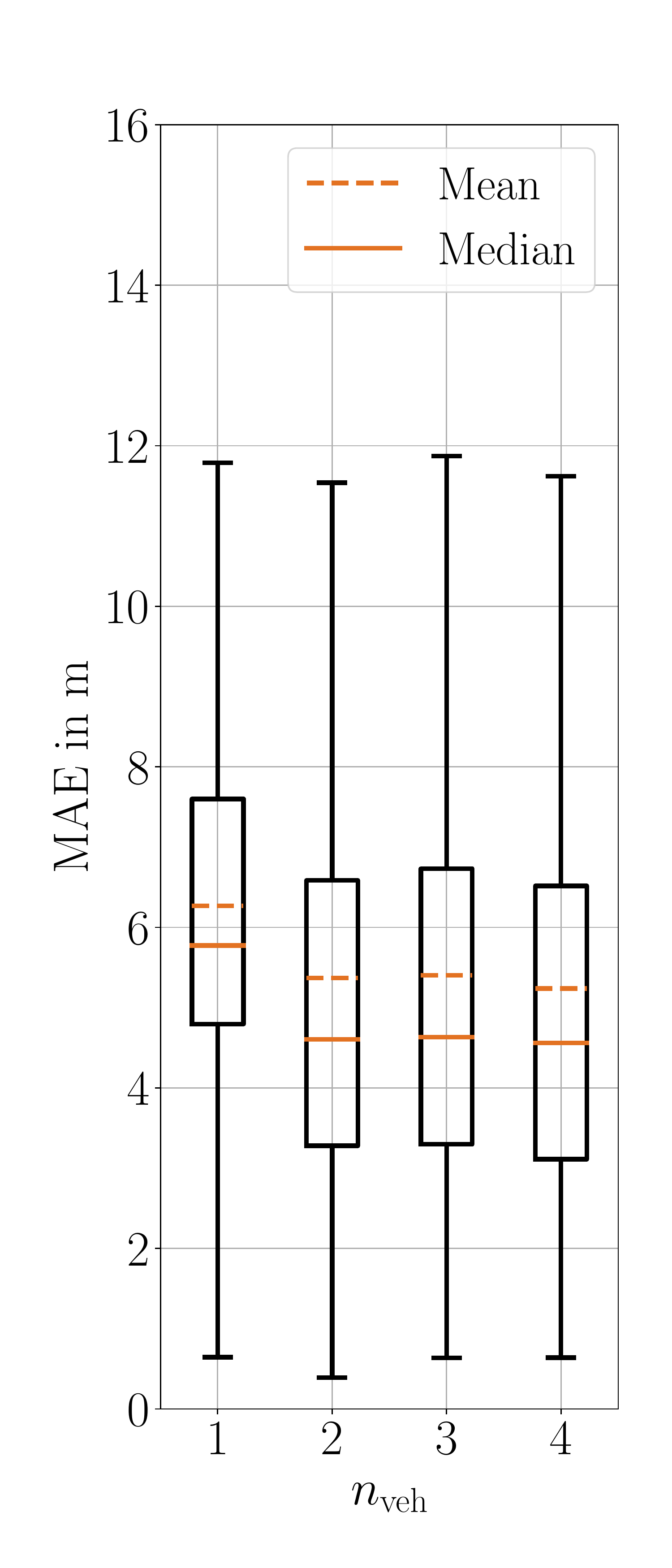}
    	\includegraphics[width=0.45\textwidth]{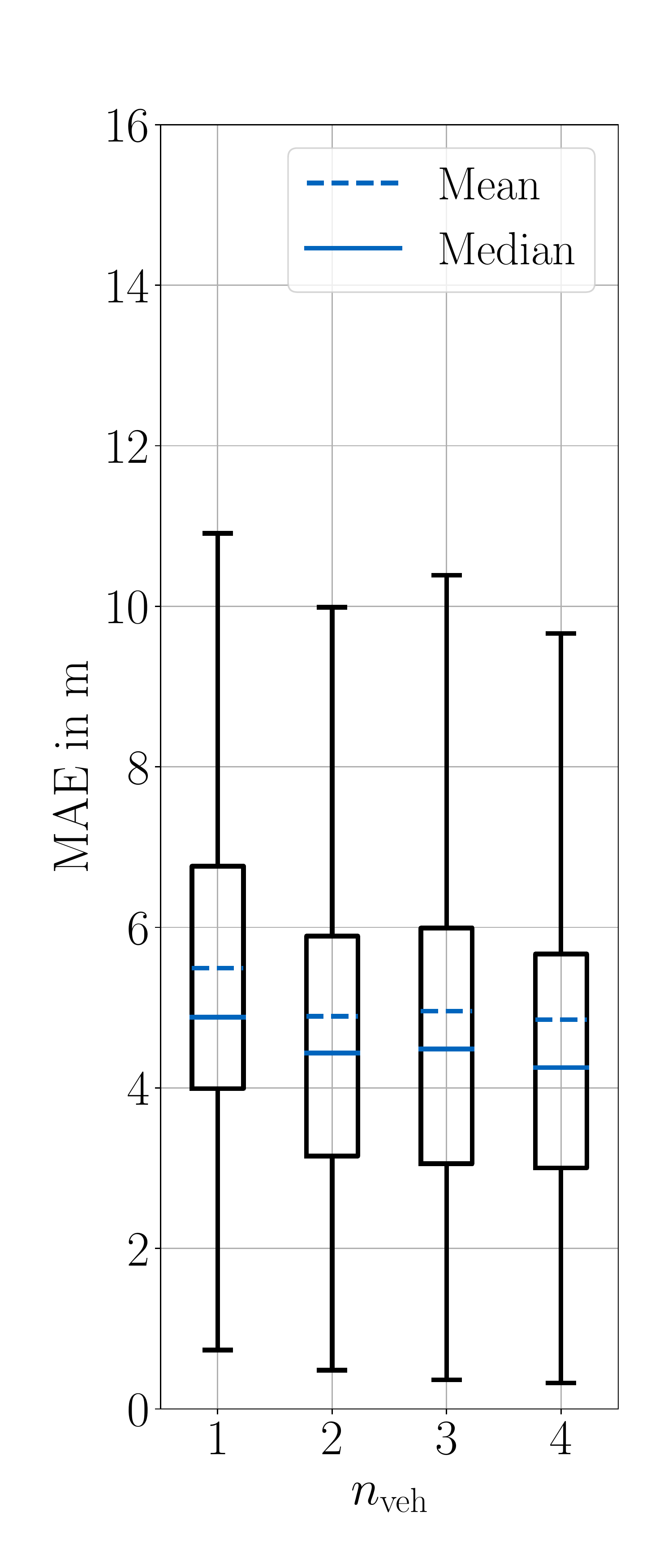}
        \caption{Number of objects.}
        \label{fig:errors_vs_nveh}
     \end{subfigure}
    \caption{The error distributions of the benchmark model (orange) and the MixNet (blue).}
\end{figure}

The error distributions with respect to the average velocity of the history are shown in Figure \ref{fig:errors_vs_velocity}. The predictions are separated into the three bins $v < \SI{30}{\metre\per\second}$,  $\SI{30}{\metre\per\second} < v < \SI{60}{\metre\per\second}$ and $\SI{60}{\metre\per\second} < v$ based on the average velocity of their histories, which are associated with a slow-speed and start scenario range, a mid-speed range and a top-speed range. As Figure \ref{fig:errors_vs_velocity} shows, the models have similar characteristics regarding their velocity dependent accuracy. The mean error is the highest at low speeds and it gradually decreases as the velocity grows. There are two main reasons for this. Firstly, most of the transient scenarios like start scenarios, which are challenging to predict, happen at lower speeds. Once the cars have reached their normal racing speed, the scenarios tend to be more steady in the longitudinal direction. Secondly, since most of the racing happens at high speeds, the majority of the training data could be acquired in this velocity range. The number of vehicles does not have a large effect on the accuracy of the predictions (Figure \ref{fig:errors_vs_nveh}). This is due to the fact that the fuzzy logic can resolve prediction conflicts accurately by deciding between right and left overtake. Moreover, maneuvers with more than two vehicles racing wheel-to-wheel at the same time, which would result in strong lateral interaction, are rare. Instead, scenarios with more than two vehicles mainly result in sequential overtaking maneuvers between two vehicles respectively.

\subsection{Robustness}

To investigate the robustness properties of both data-based approaches, we have replayed the test scenarios with extra Gaussian noise added. It should be noted that the original test data is already noisy, but its magnitude is much smaller. We carried out several experiments, all with zero-mean disturbances with different variances in the lateral and longitudinal directions. Table \ref{tab:robust_errors} reviews the MAEs of the approaches in the different test cases. As the analysis reveals, MixNet and the benchmark model are both robust against the added noise in the chosen range, although performance degradation occurs. However, it can be noticed that the MAE of the MixNet is lower in all cases. In the case of adding noise in lateral and longitudinal direction, the MAE of the benchmark model increases less than that of the MixNet in case of a standard deviation of \SI{0.5}{m} in each direction. However, a bigger standard deviation of \SI{1.0}{m} in both direction results in a significant increase in the MAE of the benchmark model by $21.2\%$. In contrast, the MixNet's relative increase in the MAE is only $13.4\%$. This observation indicates that the constraints applied to the MixNet result in the desired model behavior that the model is more robust against variations of the input.
The application of Gaussian noise in only one direction reveals that the benchmark model is less sensitive in the lateral direction. However, the relative increase in the MAE for both models is similar. The application of Gaussian noise in the longitudinal direction clearly shows a big advantage of the MixNet. The incorporation of external velocity information from the tracking module results in a significantly more robust prediction accuracy as the relative increase in the MAE is only half as big as in the case of the benchmark model.

\begin{table}[b!]
\caption{Robustness of the the benchmark model and MixNet against zero-mean Gaussian noise.}
\label{tab:robust_errors}
\begin{center}
\begin{tabular}{|c|c|c|c|c|c|}
    \hline
    \multicolumn{2}{|c|}{\textbf{Standard Deviation}} & \multicolumn{2}{c|}{\textbf{Benchmark Model}} & \multicolumn{2}{c|}{\textbf{MixNet}} \\
	\hline
	$\sigma_{\mathrm{lon}}$ in m & $\sigma_{\mathrm{lat}}$ in m & abs. MAE in m & rel. MAE in \%& abs. MAE in m & rel. MAE in \% \\
    \hline\hline
    $ 0.0 $ & $ 0.0 $ & $5.36$ & - & $4.91$ & - \\
    \hline
    $ 0.5 $ & $ 0.5 $ & $5.61$ & $+ 4.6$  & $5.23$ & $+ 6.5$ \\
    \hline
    $ 1.0 $ & $ 1.0 $ & $6.50$ & $+ 21.2$ & $5.57$ & $+ 13.4 $\\
    \hline
    $ 0.0 $ & $ 1.0 $ & $5.76$ & $+ 7.5$& $5.39$ & $+ 9.7$  \\
    \hline
    $ 1.0 $ & $ 0.0 $ & $6.09$ & $+ 13.6$ & $5.29$ & $+ 7.7$\\
  \hline
\end{tabular}
\end{center}
\end{table}

\subsection{Superpositioning weights}
\begin{figure}[h!]
	\includegraphics[width=\textwidth]{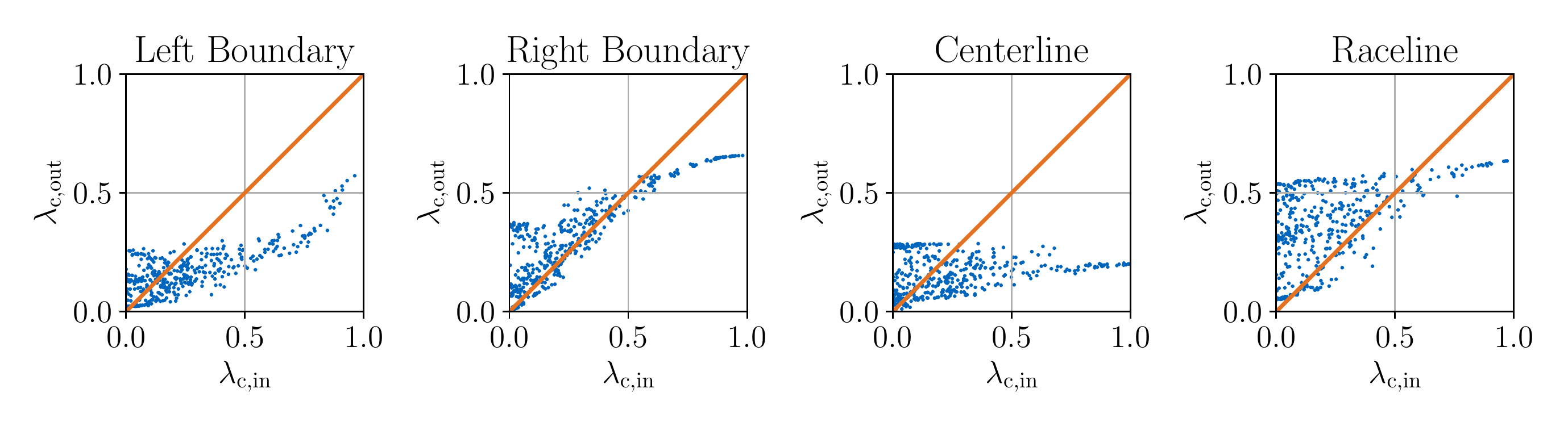}
	\caption{The relation between the inputted superpositioning weights of the synthetically generated trajectories and the outputted weights of MixNet.}
	\label{fig:weights}
\end{figure}

To investigate how consistently MixNet predicts the weights for curve superpositioning, we input synthetic history trajectories generated with random weights at the entrance of one of the turns on the racetrack. We then observe how well the outputs of MixNet match the inputs. Figure \ref{fig:weights} illustrates the input and output weights of the four base curves in scatter plots.

\setlength{\columnsep}{10pt}
\begin{wrapfigure}[27]{r}{0.5\textwidth}
	\centering 
	\includegraphics[clip, trim=1.0cm 1.0cm 1.0cm 1.1cm,width=0.5\textwidth]{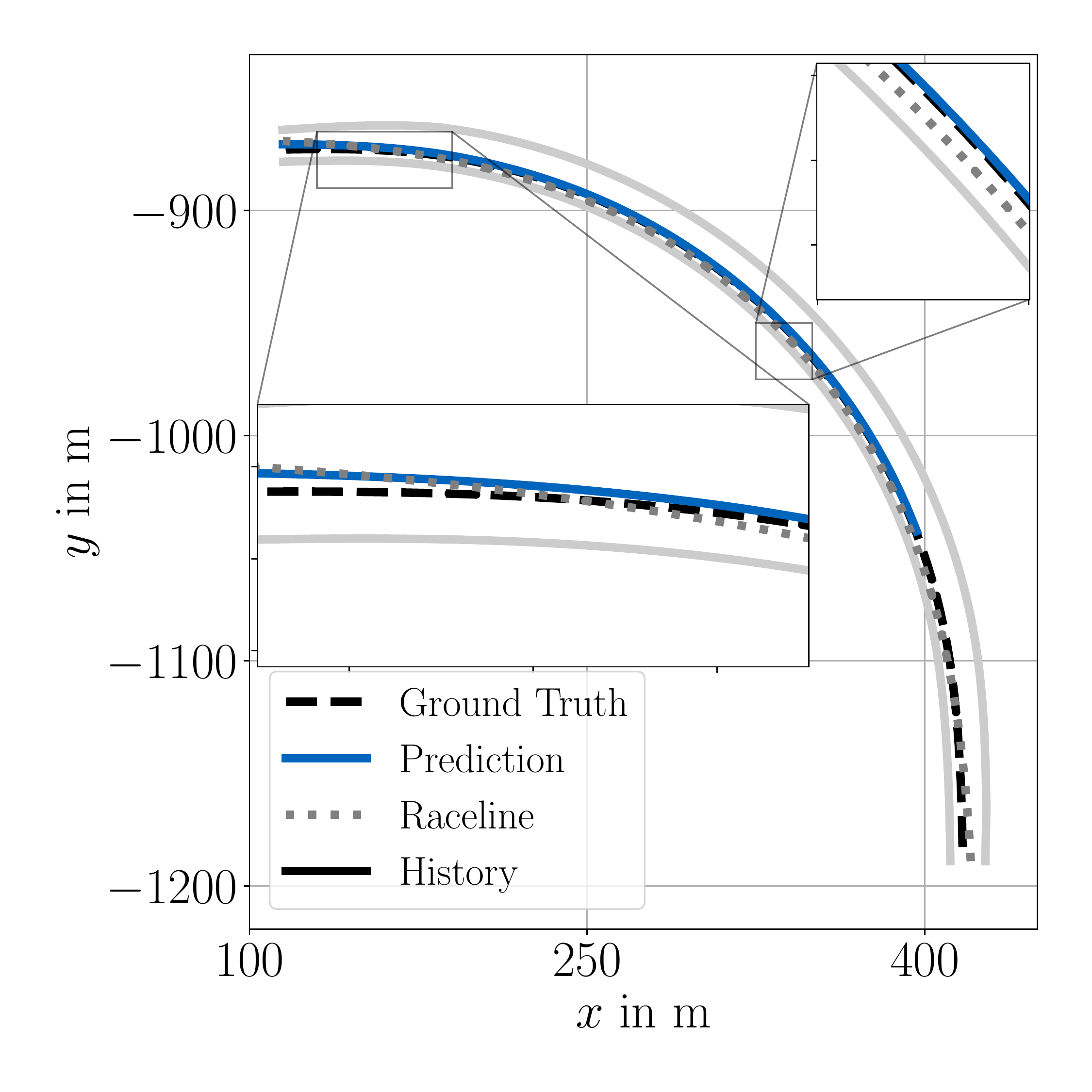}
	\caption{Exemplary scenario with overuse of the raceline in a turn.}
	\label{fig:curve_sample}
\end{wrapfigure}
Ideally, the model would output the exact same weights with which the input history is built up resulting in \SI{45}{\degree} straight lines in all of the plots. The closest to this is the figure of the raceline weights. Meanwhile, the centerline weights hardly seem to correlate with the input at all. Generally, the network seems to overuse the raceline. The reason for that is that the trajectories in the data usually have a strong raceline component. Hence, the network often assumes raceline following, although the current object might differ from the raceline. Such an example of raceline overuse can be seen in Figure \ref{fig:curve_sample}. In the turn, the prediction fits the ground truth very well close to the inner bound. However, in the turn exit, the model assumes more optimal driving and a lateral error occurs compared to the ground truth, which follows a more narrow line. The example also reveals that even though the same lateral position at a specific point results, a different combination of superposition weights influences the overall prediction path. As it can be derived from Equation \ref{eq:1} and \ref{eq:2} any point on the racetrack can be obtained through infinitely many different weighting combinations of the 4 base curves. Thus, in the shown scenario, a higher weight of the left boundary would result in the same lateral position in the turn, but a different, in this case the correct, position at the turn exit.

Even though the superposition weights differ from the inputted ones, the overall errors of the prediction are very low in these synthetic cases. The fact that the centerline is underused indicates that this base curve is indeed redundant as it lies close to the middle between left and right boundary for the major part of the track. Hence, we conduct the following analysis to prove this hypothesis and approximate the centerline as follows:
\begin{equation}
    \rho_{center}(s) \approx 0.5 \cdot \rho_{left}(s) + 0.5 \cdot \rho_{right}(s)
\end{equation}

In this case, the weights corresponding to the centerline can be redistributed and added to the left and right boundary weights without changing the overall superpositioned output as follows: 
\begin{equation}
    \lambda_{left} = \lambda_{left} + 0.5 \cdot \lambda_{center}
\end{equation}
\begin{equation}
    \lambda_{left} = \lambda_{left} + 0.5 \cdot \lambda_{center}
\end{equation}

\setlength{\columnsep}{10pt}
\begin{wrapfigure}[14]{r}{0.5\textwidth}
	\centering 
	\includegraphics[clip, trim=0.5cm 1.0cm 0.0cm 0.4cm,width=0.5\textwidth]{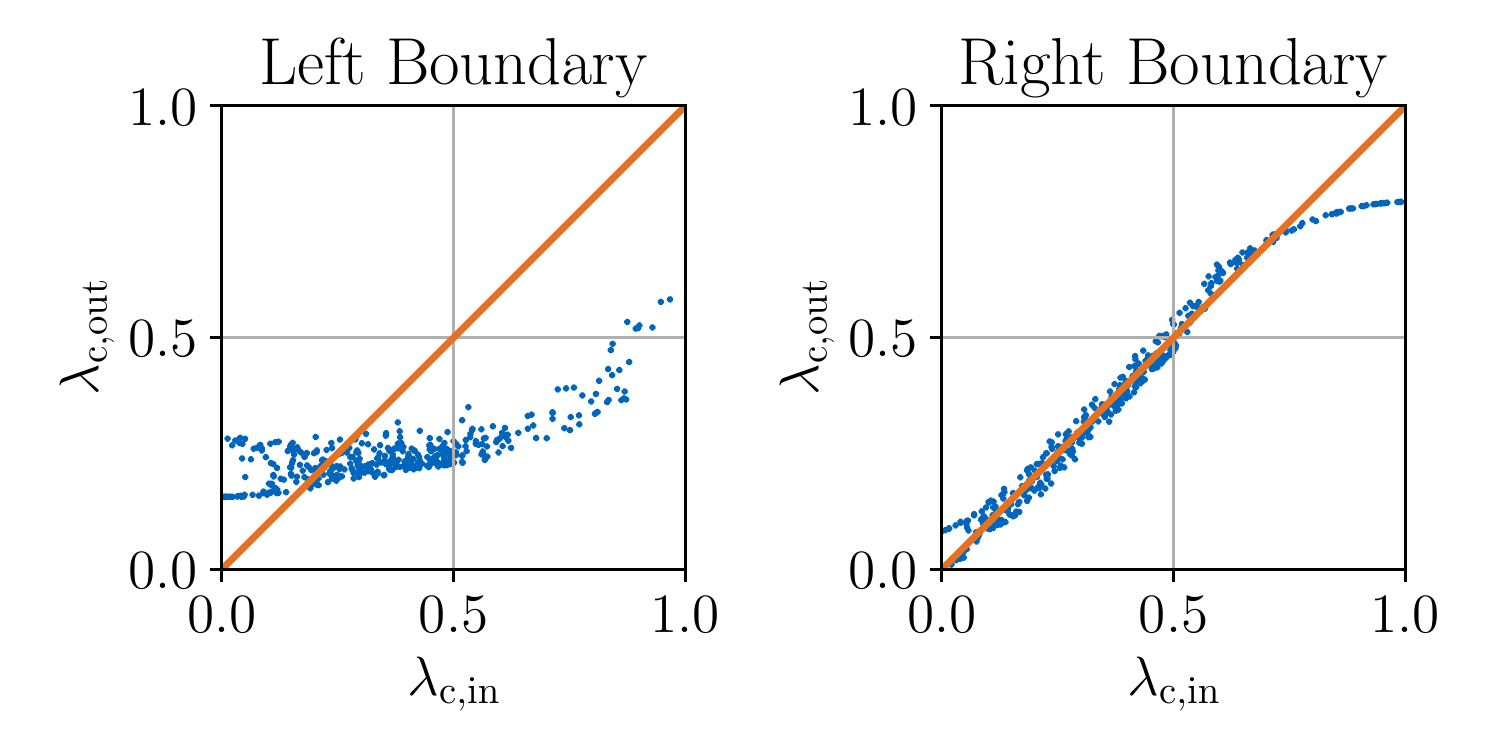}
	\caption{The relation between the input superpositioning weights of the synthetically generated trajectories and the outputted weights of MixNet if one redistributes the weights of the centerline to the right and left boundaries.}
	\label{fig:weights_without_center}
\end{wrapfigure}
If we plot the left and right boundary weights again, we get the input-output weight relationships illustrated in Figure \ref{fig:weights_without_center}.
Here, the weights for the right boundary already match very well the desired linear figure. The left boundary is still mostly substituted by over weighting the raceline. From this we conclude that the superpositioning weights produced by the network, although they do not match exactly the input weights, which is also not expected, due to the redundancy of the base curves, are consequent and reasonable.

\subsection{Computation time}
The average computation times for predicting four vehicles on a single core of an Intel i7-4720HQ 2.6 GHz CPU for MixNet is \SI{9}{\milli\second} and \SI{15}{\milli\second} for the benchmark model. The models have encoders of identical sizes, but the LSTM-decoder of the benchmark model takes longer to execute due to iterative calculation of the vehicles' future trajectory, which is the reason for the higher calculation time. Even on a CPU-single core the computation time is low enough that the model can be applied in a full software stack without access to the GPU, which would reduce the computation time significantly.


\section{Future Work} \label{section:future_work}

An interesting future research direction could be to determine a richer set of base curves for superpositioning. Including for example base curves which cover lateral motion on the track could result in even higher expressiveness. Additionally, the flexibility of the approach could be enhanced by outputting multiple weights along the path instead of one constant set of weight parameters. Thus, the trade off between accurate prediction towards the end of the prediction horizon and a low lateral offset between the current object position and the start of the predicted trajectory could be solved. However, both future directions require more data and add complexity to the training process. Besides that the robustness for real-world application has to be reviewed in case of the more flexible network with varying superpositioning weights.

Moreover, interaction-awareness could be improved by directly incorporating surrounding objects in the network input. The state of the art offers various methods such as grid-based methods or GNNs to model interactions. Our presented rule-base approach is robust and explainable and performances well with a low number of cars ($< 4$). However, the scalability of the approach is limited because the sequential application of the rule-base overtaking decision results in an increasing number of prediction collisions, when aplied to a higher number of vehicles.

Another idea that would be worth investigating is how the knowledge of MixNet could be generalized to previously unseen tracks. Although the predicted path can be constrained by choosing the respective base lines to lie inside them, the learned scenario understanding depends on the applied base curves. This can be explained by the fact that the network itself does not know the base curves, hence the forecasts on another racetrack, which inherently comes with different base curves, would not result in meaningful forecasts. This shortcoming, however, could be remedied if the base curves would also be input to the model. Another option enhance the generalization capabilities is to build a modular network with a a backbone network with a general motion encoding, which is trained on a large scale dataset and a track specific head.

Lastly, we would like to draw the attention to an additional feature of MixNet that could be further explored. Namely, that the path predicted does not only exists in the \textit{future}. Superpositioning the base curves according to the predicted weights results in a path that covers a complete lap on the track. Hence, it also exists behind the current position of the vehicle. This means, that the predicted path can directly be compared to the history trajectory of an object. This can be useful for example in obtaining a confidence value for each forecast. It is somewhat similar to what we have exploited when we filtered and overrode forecasts which were identified as potentially incorrect at the beginning of the horizon. However, instead of using only the most resent position of the vehicle, one could use all the historical states and hence get a picture of how well the predicted path fits the observed history of the car.


\section{Conclusion} \label{section:conclusion}

Autonomous driving has been an emerging research field in the past few decades. To fuel innovation, autonomous car races have been and are being organized. The first high-speed wheel-to-wheel competition was the Indy Autonomous Challenge. This paper described the motion prediction algorithm of the team TUM Autonomous Motorsport developed for this race.

Our solution is an encoder-decoder neural network architecture, called MixNet, which carries out the motion prediction task in a structured manner. The encoder part of the network creates a latent summary from the history trajectories and some relevant map information. The decoder then creates a path by superpositioning predefined base curves with weights predicted by the network. It also generates a piece-wise linear velocity profile, according to which the path is resampled to reach the output trajectory. By this, the approach is capable to combine a comprehensive scenario understanding of the RNN-encoder network and constrained, thus robust, prediction output superposed by kinematically feasible base curves.

The main contribution of this work is the development of a structured deep neural motion prediction algorithm that allows giving some quality guarantees about its output. The predicted trajectories are guaranteed to be smooth and lie inside the racetrack. Meanwhile these guarantees can be given, thanks to the highly restricted prediction set and the method is still flexible enough to produce accurate predictions in the majority of the cases. Also, since the velocity profile is predicted separately, it is possible to incorporate velocity information from other sources, such as the object tracking module's state estimator filters. Furthermore, it is possible to detect predictions that are probably inaccurate at the beginning of the horizon. These predictions could easily lead to dangerous behavior, hence filtering and overriding them are crucial in a safety-critical application like autonomous racing. Finally, a rule-based overtaking logic allows to resolve collisions between predicted trajectories. The algorithm is real-time capable on a single CPU core and its applicability in the overall software stack of the team is proven. The whole software module is available open source including the training and test data, a ROS2-launch configuration and a build file to create a Docker image to run the module containerized.

To demonstrate the performance of the method, we compared its accuracy to that of an unrestricted LSTM-based encoder-decoder architecture. The results underline that the highly restricted prediction set of MixNet does not cause large performance degradation. Contrarily, the lateral errors were almost identical in both cases and since MixNet is capable of incorporating velocity information from the tracking module, it even outperformed the benchmark model in the overall accuracy. The model also shows great robustness to noised inputs. Finally, we investigated how consequently the network produces the superpositioning weights.

\section*{Contributions}
Phillip Karle as the first author initiated the idea of this paper and contributed essentially to its conception, implementation and content. Ferenc Török developed the concept of MixNet, contributed the data generation, training and evaluation procedure and contributed to the writing of the paper. Maximilian Geisslinger contributed to the conception and implementation of this research and revision of the research article. Markus Lienkamp made an essential contribution to the conception of the research project. He revised the paper critically for important intellectual content. He gave final approval of the version to be published and agrees to all aspects of the work. As a guarantor, he accepts the responsibility for the overall integrity of the paper.

\bibliographystyle{apalike}
\bibliography{references}

\end{document}